\documentclass{article} 
\usepackage{iclr2021_conference,times}

\usepackage{amsmath,amsfonts,bm}

\def\eqref#1{equation~\ref{#1}}

\def\1{\bm{1}}

\DeclareMathAlphabet{\mathsfit}{\encodingdefault}{\sfdefault}{m}{sl}
\SetMathAlphabet{\mathsfit}{bold}{\encodingdefault}{\sfdefault}{bx}{n}

\usepackage{hyperref}
\usepackage{url}
\usepackage{graphicx}
\usepackage{caption}
\usepackage{subcaption}
\usepackage{float}
\usepackage{mathabx}

\hypersetup{
    colorlinks=true,
    linkcolor=red,
    citecolor=blue,
    filecolor=magenta,
    urlcolor=cyan,
}

\newcommand{\mvec}[1]{\mathbf{#1}}
\newcommand{\rstar}{\mvec{r}^*}

\title{The Emergence of Abstract and Episodic Neurons in Episodic Meta-RL}

\author{Badr AlKhamissi \thanks{\,Corresponding author} \\
Sony CSL \\
Tokyo, Japan \\
\texttt{badr [at] khamissi.com} \\
\And
Muhammad ElNokrashy \\
Microsoft EGDC \\
Cairo, Egypt \\
\texttt{muelnokr [at] microsoft.com} \\
\AND
Michael Spranger \\
Sony CSL \\
Tokyo, Japan \\
\texttt{michael.spranger [at] sony.com}
}

\iclrfinalcopy 
\begin{document}

\maketitle

\begin{abstract}
In this work, we analyze the reinstatement mechanism introduced by \cite{ritter19been} to reveal two classes of neurons that emerge in the agent's working memory (an epLSTM cell) when trained using episodic meta-RL on an episodic variant of the Harlow visual fixation task. Specifically, \emph{Abstract} neurons encode knowledge shared across tasks, while \emph{Episodic} neurons carry information relevant for a specific episode's task.

\end{abstract}

\section{Introduction}

Starting as a method to study animal conditioning in psychology \citep{pavlov1927, rescorla72}, \emph{Reinforcement Learning} (RL) has become an efficient way to train artificial agents in solving complex tasks such as Go or StarCraft \citep{alphago16,alphastar}. Despite such successes important problems remain. State-of-the-art RL algorithms require enormous amount of training data and do not easily adapt to new tasks. 

One research strand trying to address these issues is  \emph{meta-reinforcement learning} (meta-RL) \citep{wang16l2rl} - in which agents have to learn to deal with a number of different tasks. Typically, in this work recurrent neural networks - specifically LSTMs - are used to learn representations that encode an RL algorithm. \cite{ritter19been} proposed to extend these LSTMs with neural memory - so the agents are able to remember and retrieve knowledge gained over past tasks when re-encountering them. The memory proposed by \cite{ritter19been} relies on a gating mechanism that decides which memory activations are retrieved and reinstated into the LSTM. This gating mechanism is trained as part of the overall optimization problem and constitutes a key artifact of learning.

We study how the gating mechanism interacts with LSTM neurons, and show that they can be roughly categorized. We identify two classes of neurons: \emph{episodic} and \emph{abstract} neurons - that differ in their characteristics w.r.t how information is restored as well as their impact on the performance of the system. \emph{Abstract neurons} encode structural task knowledge relevant across different episodes, while \emph{Episodic neurons} carry episode-specific environmental information (potentially reoccurring in later episodes).

This paper proceeds as follows: (1) we introduce a simplified version of the Harlow Task (a standard Meta-RL environment) with episodic cues, (2) we introduce the model implementation, followed by (3) an analysis, definition and tests for abstract and episodic neurons. Lastly, we briefly discuss our results in the wider context of meta-RL.

\section{Methods}
\subsection{Task Formulation}
\label{subsec:problem_formulation}

\begin{figure}[h]
\centering
\begin{subfigure}{.2\textwidth}
    \begin{subfigure}{1\textwidth}
      \centering
      \includegraphics[width=1\linewidth]{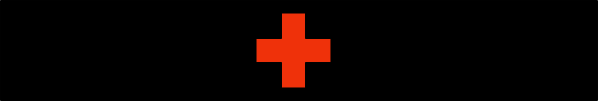}
    \caption{} 
    \label{fig:harlow_1d_a}
    \end{subfigure} 
    \begin{subfigure}{1\textwidth}
      \vspace{4mm}
      \centering
      \includegraphics[width=1\linewidth]{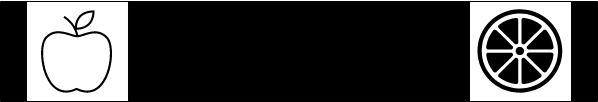}
    \caption{} 
    \label{fig:harlow_1d_b}
    \end{subfigure}
\end{subfigure}%
\begin{subfigure}{.15\textwidth}
    \centering
    \includegraphics[width=0.8\linewidth]{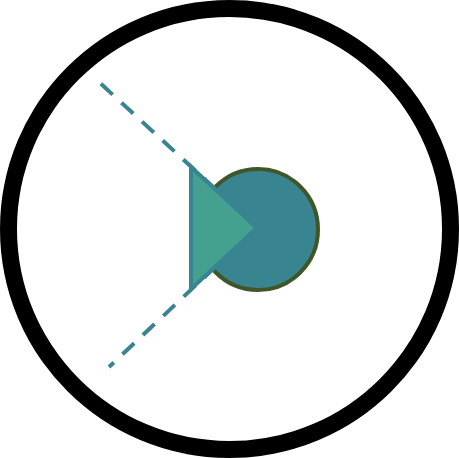}
    \caption{} 
    \label{fig:harlow_1d_c}
\end{subfigure}%
\begin{subfigure}{0.31\textwidth}
    \centering
    \includegraphics[width=1\linewidth]{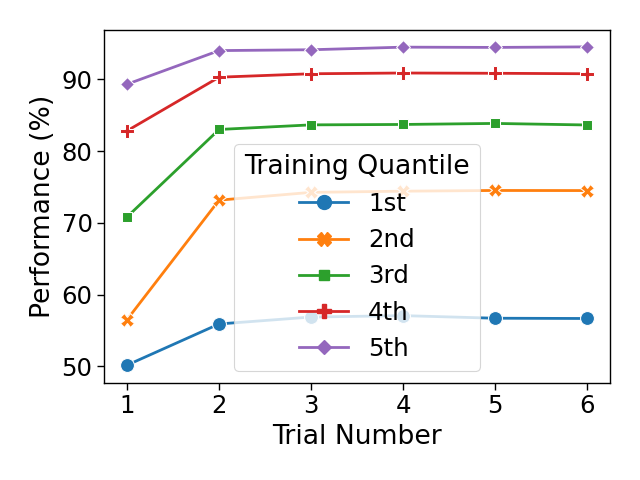}
    \caption{} 
    \label{fig:training_per_quantile}
\end{subfigure}%
\begin{subfigure}{0.31\textwidth}
    \centering
    \includegraphics[width=1\linewidth]{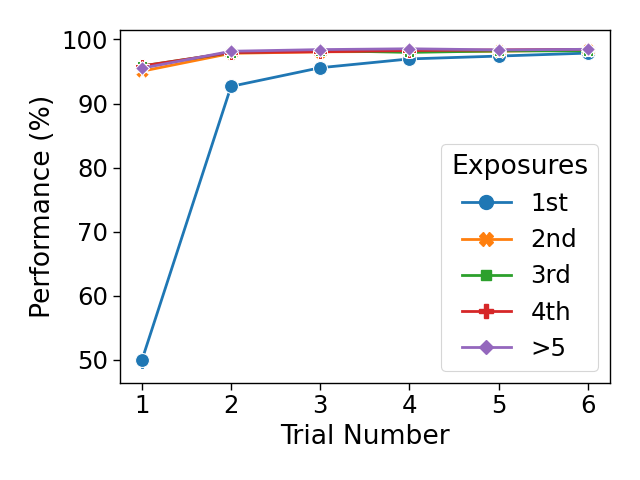}
    \caption{} 
    \label{fig:testing_per_exposure}
\end{subfigure}
\caption{Illustration of the 1D Symbolic Harlow task and its training and testing performance.
\textbf{(a)} Fixation cross at the center of agent's receptive field.
\textbf{(b)} Objects placed in agent's receptive field upon fixation.
\textbf{(c)} Top-down view of agent in the environment.
\textbf{(d)} Average training performance at each trial number, per training quantile.
\textbf{(e)} Testing performance at each trial number, per number of exposures to a specific task.}
\label{fig:harlow_task}
\end{figure}

Let $M_i \in \mathcal{D}$ be a distribution of tasks each characterized as a Markov Decision Process (MDP): $M_i = (\mathcal{S}, \mathcal{A}, \mathcal{T}_i, \mathcal{R}_i)$. The agent learns over a sequence of MDPs by taking an action $a \in \mathcal{A}$ to transition from state $s$ to $s'$ (where $s, s' \in \mathcal{S}$) and receiving a scalar reward $r$, using some transition probability distribution $\mathcal{T} : \mathcal{S} \times \mathcal{A} \rightarrow \mathcal{S}$ and reward function $\mathcal{R} : \mathcal{S} \times \mathcal{A} \rightarrow \mathbb{R}$. To introduce the concept of identifiably reoccurring tasks, \cite{ritter19been} extend the previous formulation by associating a context $k_i$ with each $M_i$, sampled as $(M_i, k_i) \sim \mathcal{D}$ uniformly with replacement. With each new task, the agent can use the context $k_i$ to identify if the task had been seen before, and hence leverage previously discovered policies to avoid redundant exploration.

\paragraph{The (One-dimensional) Symbolic Episodic Harlow Task}
To study and analyze episodic meta-RL agents, we develop a simplified symbolic version with exact parallels to the task structure of the Harlow visual fixation task found in the PsychLab environment \citep{psychlab} but which factors out the visual and spatial modeling of the environment. Further details can be found in Appendix \ref{sec:task}.

\subsection{Model Description}
The agent is trained\footnote{All experiments were done on a single Nvidia GeForce RTX 2080Ti.} using the Advantage Actor-Critic (A2C) RL algorithm on a single thread \citep{mnih15humanlevel}. The architecture follows that of the LSTM A3C model from \cite{wang16l2rl} but uses an epLSTM instead. The encoder is a stack of $2$ affine layers with $64$ and $128$ units, respectively, and a ReLU non-linearity in-between. The epLSTM takes a concatenation of: \textbf{(a)} The encoding of the receptive field, \textbf{(b)} the reward at $t-1$, and \textbf{(c)} the action at $t-1$. The epLSTM is a one layer LSTM with $256$\footnote{Smaller models showed the same results as will follow, but took longer to converge.} hidden units plus the reinstatement mechanism. The memory module maps the context associated with the current task $k_i$ (as key) to the cell state $\mvec{c}_{T}$ (as value) at the end of each episode (time $T$). This memory is updated at the same key each time the task $M_i$ reoccurs. The experiment is repeated $50$ times with different initializations. We analyze the top $30$ models (filtered by a threshold on the reward calculated on $100$ randomly generated episodes). Each instance is trained for $25,000$ episodes and tested for $1,000$ episodes (with different objects). The code is made open-source\footnote{\url{https://github.com/BKHMSI/emrl-neuron-emergence}}.

\subsection{Reinstatement Mechanism}
We use the reinstatement mechanism from \citep{ritter19been}---defined as: 

\begin{equation}
     \label{eq:rs_mech}
     \mvec{r}_t =
     \sigma\left(\mvec{W}_{rx}
            \mvec{x}_t
        + \mvec{W}_{rh}
            \mvec{h}_{t-1} + \mvec{b}_r\right)
\end{equation}

where $\mvec{r}_t$ controls the flow of information from the retrieved memory $\mvec{m}_t$ into the epLSTM cell state:

\begin{equation}
    \label{eq:ep-cell-state}
        \mvec{c}_t
            =
            \mvec{i}_t \odot \mvec{\tilde{c}}_{t}
            +
            \mvec{f}_t \odot \mvec{c}_{t-1}
            +
            \underline{\mvec{r}_t \odot \tanh\left(\mvec{m}_t\right)}
\end{equation}

The vector $\mvec{c}_t$ can be seen as encoding the agent's working memory state up to time $t$. Therefore, at the end of each episode the agent commits the accumulated knowledge learned about the current MDP $M_i$ to an external long-term memory module with the associated context vector $k_i$ as key. In a later episode where $M_i$ reoccurs, the context $k_i$ is used to query the memory to retrieve the corresponding cell state as $\mvec{m}_t$ (equals the zero vector if $M_i$ is novel). The retrieval occurs when the agent first observes the pair of objects (after fixation at the first trial). The vector $\mvec{m}_t$ is then interpolated into the current working memory using the reinstatement gate $\mvec{r}_t$.

\section{Analysis}
\label{sec:analysis}

\subsection{Task Performance}
Figure \ref{fig:training_per_quantile} shows the performance over successive stages of training as a function of the trial number. The performance on the first trial improves and is not stuck at random as in the classical Harlow experiment because the agent is able to reinstate relevant information when it re-encounters a specific task. Figure \ref{fig:testing_per_exposure} shows the testing performance as a function of the number of times an agent is exposed to a particular task. It can be seen that when a task reoccurs the agent immediately identifies it and is able to solve it from the first trial.

\subsection{Recurrent Neurons and the Reinstatement Gate}
\begin{figure}[H]
\centering
\begin{subfigure}{0.33\textwidth}
    \centering
    \includegraphics[width=1.0\linewidth]{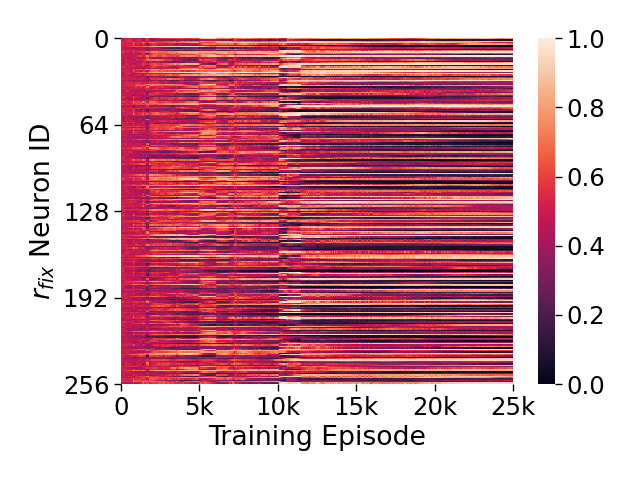}
    \caption{}
    \label{fig:rt_history}
\end{subfigure}%
\begin{subfigure}{.33\textwidth}
    \centering
    \includegraphics[width=1.0\linewidth]{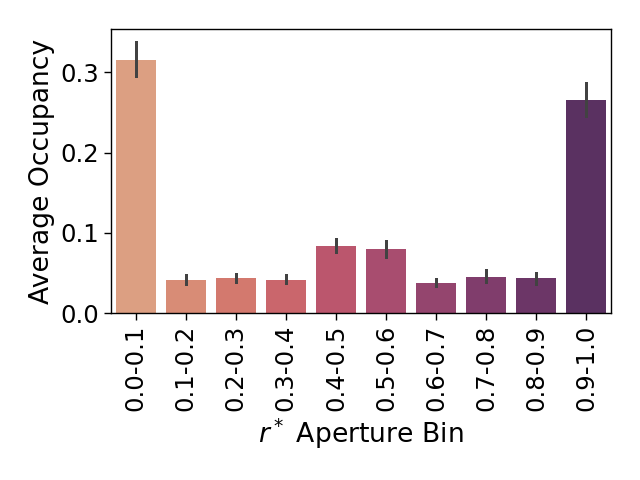}
    \caption{}
    \label{fig:rt_stats}
\end{subfigure}%
\begin{subfigure}{0.33\textwidth}
    \centering
    \includegraphics[width=1.0\linewidth]{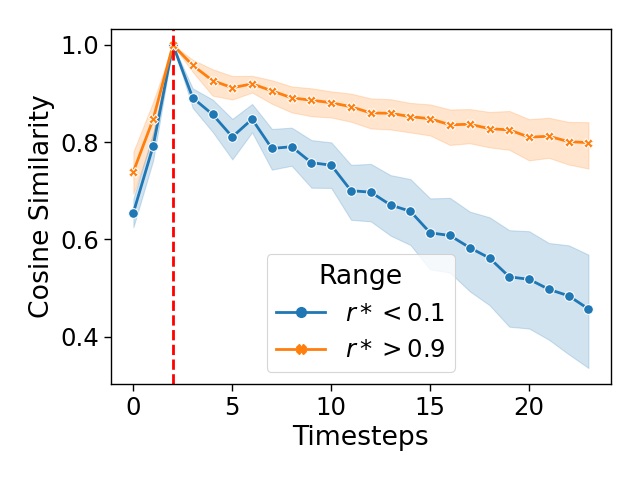}
    \caption{}
    \label{fig:cosine_consistency}
\end{subfigure}
\caption{
\textbf{(a)} The values of $\mvec{r}_\text{fix}[j]$ for each neuron $j$ across training episodes.
\textbf{(b)} Average percentage of neurons in $\rstar$ that appear within a certain bin of ``openness" during testing across $30$ different seeds.
\textbf{(c)} Average vector similarity between $\mvec{c}_\text{fix}$ at the first fixation in an episode (vertical line), and $\mvec{c}_{t}$ at every other step.}
\label{fig:analysis_openness}
\end{figure}

Noting that $\mvec{r}_t[j] \in (0, 1)$ (by the $\sigma$ function) modulates the reinstatement of individual neurons from $\mvec{m}_t$, we may interpret $\mvec{r}_t[j]$ as the importance of neuron $j$ for recurring episodic information.

Figure \ref{fig:rt_history} shows the values of $\mvec{r}[j]$ at fixation (as heat) for each neuron $j$ across training episodes. Notice that the $\mvec{r}[j]$ activations converge to stable values as training progresses, independent of changes in input or hidden state among the different episodes. We calculate from the last $1000$ training episodes a static value for $\mvec{r}_\text{fix}$ ($\mvec{r}_t$ at $t=\text{fixation}$) and use it for all further gate analysis. Formally ($e$ is the episode index and $N_e$ is the training episode count):

\begin{equation}
    \mvec{r}^* =
        \operatorname*{mean}_e \left\{
            \mvec{r}_\text{fix}^e
            : N_e - 1000 < e \le N_e
        \right\}
\end{equation}

Figure \ref{fig:rt_stats} is a histogram of $\rstar[j]$ values. About $25\%$ of the neurons have become biased to be open ($\rstar[j] \ge 0.9$), while ${\sim}30\%$ are biased to be closed ($\rstar[j] < 0.1$). Figure \ref{fig:cosine_consistency} shows the cosine similarity of certain regions (indicated by hue) of $\mvec{c}_t$ to $\mvec{c}_\text{fix}$ (averaged over $1000$ testing episodes). This suggests that some specific neurons consistently hold the information needed across episodes to identify the winning object, and so change values less often within each episode (see Figure \ref{fig:analysis_masking}).

\subsection{Testing for Abstract and Episodic Neurons}

\begin{figure}[h]
    \centering
\begin{subfigure}{0.4\textwidth}
    \centering
    \includegraphics[width=1\linewidth]{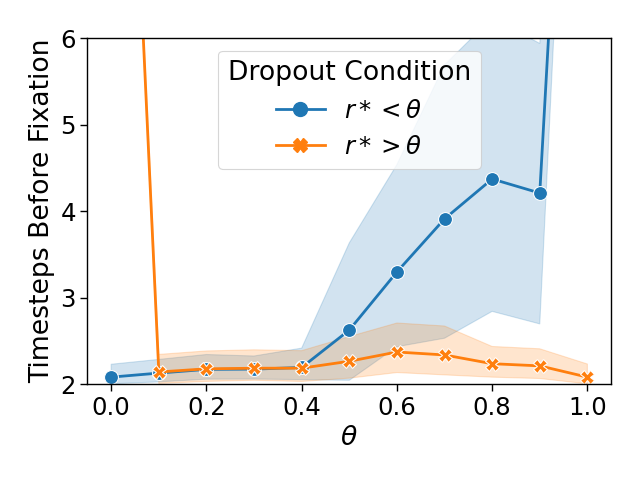}
    \caption{Time steps}
    \label{fig:perf_v_steps}
\end{subfigure}%
\begin{subfigure}{0.4\textwidth}
    \centering
    \includegraphics[width=1\linewidth]{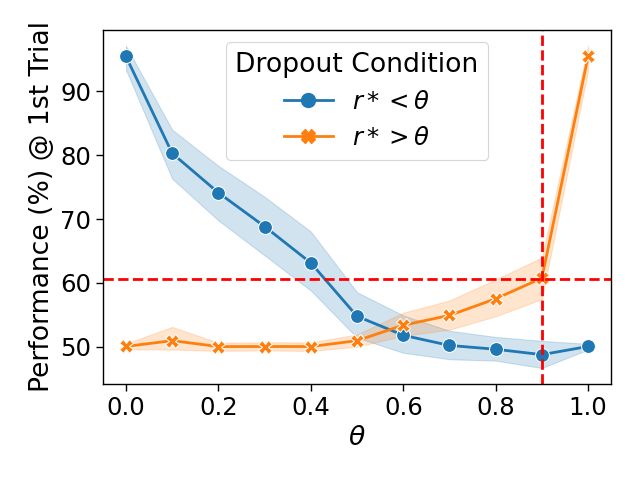}
    \caption{Performance}
    \label{fig:perf_per_mask_gradual}
\end{subfigure}%
\caption{
\textbf{(a)} Average time-steps before fixating when dropping \emph{episodic} ($\mvec{r}^*\ge\theta$) or \emph{abstract} ($\mvec{r}^*<\theta$) neurons at $\theta$.
\textbf{(b)} Average first trial performance at thresholds $\theta$. Note the $30\%$ regression when dropping \emph{episodic} neurons at $\theta=0.9$ (dashed lines).
}
\label{fig:analysis_masking}
\end{figure}

To clarify the roles of individual neurons in $\mvec{c}_t$, we test while gradually masking out the neurons $\mvec{c}_t[j]$ based on $\left\{j : \rstar[j] \lessgtr \theta\right\}$ then analyze the behavioral change in two signals: \textbf{(a)} number of steps before fixation, and \textbf{(b)} first trial performance.

Figure \ref{fig:perf_v_steps} shows the average time-steps before fixation when zeroing out $\mvec{c}_t$ based on the upper/\textit{episodic} (orange) or lower/\textit{abstract} (blue) regions of $\rstar$. Dropping more of the ``abstract" region leads to worse performance (more steps to fixation). Performance remains largely stable as we drop ``episodic" neurons first, up until before the extreme where all neurons are dropped ($\theta = 0.0$).
    
Figure \ref{fig:perf_per_mask_gradual} shows objective performance (choosing the rewarding object) at the first trial during testing where the MDP $M_i$ has occurred before. Dropping neurons from the ``abstract" region (blue curve) shows a smooth drop in performance, while dropping ``episodic" neurons shows a steep drop from as early as $\rstar > 0.9$, suggesting a strong correlation between the neurons where $\rstar[j] > \theta$ (for some reasonable $\theta$) and task performance on object re-occurrence. This seems to indicate that this region holds most of the object-based reward information.

\section{Discussion}

In this work, we have shown the existence of two classes of neurons that emerge in the memory reinstatement-based episodic meta-RL paradigm. Each neuron may belong to a class that encodes episodic information, or a class that encodes abstract knowledge that is shared across episodes. This finding implies that one does not need to store the whole cell-state when committing it to the long-term memory module since only a fraction of the activations are actually going to be reinstated. Therefore, one optimization method is to store a sparse representation of $c_T$ while storing the required indices only once. In the case of the experiments conducted in this paper, this method can save up to $75\%$ of the storage cost for the memory module while maintaining close to optimal performance after deployment once $\rstar$ is computed.

\cite{wang18pfc} had shown that the meta-RL framework has direct connections with structures and functions in the brain. Specifically, they conceptualize the prefrontal cortex (PFC) along with the subcortical structures to which it connects as forming a homogeneous recurrent neural network that is trained using striatal dopamine reward prediction error signals. Inline with this theory and the work presented in this paper, previous work have shown that the PFC contain single neurons that encodes abstract rules \citep{wallis01abstract}. Future work may extend the analysis to different episodic tasks, and utilize the findings for incorporating stronger inductive biases. We hope this work meaningfully furthers the sharing of insights between the neuroscience and machine learning fields. 




\bibliography{references}
\bibliographystyle{iclr2021_conference}

\appendix
\section{The Symbolic Episodic Harlow Task}
\label{sec:task}

The task consists of a one-dimensional circular state-space with $16$ discrete cells, of which the agent can observe $8$ cells at any time step (see Figure \ref{fig:harlow_task} for an illustration of the task). It starts with a central fixation cross placed in the initial observable space (i.e. receptive field) of the agent; similar to the PsychLab version. The agent can then select one of two actions $|\mathcal{A}|=2$: move one cell to the \textit{left} or to the \textit{right}. After the fixation cross appears in the center of the agent's receptive field, it is removed and two objects are introduced to the left and right of the center, one of which is randomly assigned to be the rewarding object throughout the episode. The objects are uniformly sampled from a distribution of $n=100$ objects split into $80$ for training and $20$ for testing, resulting in $n(n-1) = 9,900$ possible combinations of object-reward pairs, because in each task either of the object pair may be rewarding. The agent must then choose one of the objects by orienting it towards the center of its receptive field. Following that, the fixation cross reappears initiating the next trial.

Similar to the PsychLab version, one episode consists of $6$ trials, but here it is terminated after a maximum of $120$ total time steps. We use the same reward values as used in \cite{wang18pfc}: $1$ and $-1$ for the rewarding and non-rewarding objects, respectively, and $0.2$ for arriving at the fixation cross. The episodic structure comes from the fact that the same objects with their associated rewards can be sampled more than once. In order for the agent to identify the current task, a context vector is randomly generated the first time the agent encounters a particular task. This creates a unique mapping between each possible MDP and its corresponding context. 

\end{document}